\documentclass[10pt,twocolumn,letterpaper]{article}

\usepackage[pagenumbers]{iccv} 

\definecolor{iccvblue}{rgb}{0.21,0.49,0.74}
\usepackage[pagebackref,breaklinks,colorlinks,allcolors=iccvblue]{hyperref}

\usepackage{multirow}
\usepackage{rotating}
\usepackage{adjustbox}
\usepackage{bbding}
\usepackage{algorithm}
\usepackage{listings}

\usepackage{etoolbox}
\makeatletter
\AfterEndEnvironment{algorithm}{\let\@algcomment\relax}
\AtEndEnvironment{algorithm}{\kern2pt\hrule\relax\vskip3pt\@algcomment}
\let\@algcomment\relax
\newcommand\algcomment[1]{\def\@algcomment{\footnotesize#1}}
\renewcommand\fs@ruled{\def\@fs@cfont{\bfseries}\let\@fs@capt\floatc@ruled
  \def\@fs@pre{\hrule height.8pt depth0pt \kern2pt}%
  \def\@fs@post{}%
  \def\@fs@mid{\kern2pt\hrule\kern2pt}%
  \let\@fs@iftopcapt\iftrue}
\makeatother

\def\paperID{1961} 
\def\confName{ICCV}
\def\confYear{2025}

\title{EmbodiedPlace: Learning Mixture-of-Features with Embodied Constraints \\ for Visual Place Recognition}

\author{Bingxi Liu$^{1,2}$, Calvin Chen$^{3}$, Shiyi Guo$^{4}$, Yihong Wu$^{5}$, Jinqiang Cui$^{2, *}$, Hong Zhang$^{1,}$\thanks{ Corresponding author.}\\
$^{1}$Southern University of Science and Technology, Shenzhen, China.\\
$^{2}$Peng Cheng Laboratory, Shenzhen, China.\\
$^{3}$University of Cambridge, Cambridge, United Kingdom.\\
$^{4}$Northeastern University, China.\\
$^{5}$MAIS, Institution of Automation, China Academic of Sciences, Beijing, China.\\
{\tt\small liubx@pcl.ac.cn, hzhang@sustech.edu.cn}}

\begin{document}
\maketitle
\begin{abstract}
Visual Place Recognition (VPR) is a scene-oriented image retrieval problem in computer vision in which re-ranking based on local features is commonly employed to improve performance. In robotics, VPR is also referred to as Loop Closure Detection, which emphasizes spatial-temporal verification within a sequence. However, designing local features specifically for VPR is impractical, and relying on motion sequences imposes limitations. Inspired by these observations, we propose a novel, simple re-ranking method that refines global features through a Mixture-of-Features (MoF) approach under embodied constraints. First, we analyze the practical feasibility of embodied constraints in VPR and categorize them according to existing datasets, which include GPS tags, sequential timestamps, local feature matching, and self-similarity matrices. We then propose a learning-based MoF weight-computation approach, utilizing a multi-metric loss function. Experiments demonstrate that our method improves the state-of-the-art (SOTA) performance on public datasets with minimal additional computational overhead. For instance, with only 25 KB of additional parameters and a processing time of 10 microseconds per frame, our method achieves a 0.9\% improvement over a DINOv2-based baseline performance on the Pitts-30k test set.
\end{abstract}    
\section{Introduction}
\label{sec:introduction}
Visual Place Recognition (VPR) involves identifying the image most similar to a query image from a large-scale geographic image database \citep{dvg}. It is a fundamental task in various applications, such as autonomous driving \cite{robotcar} and augmented reality \cite{long_vl}. From the era of hand-crafted methods \cite{bow} to the deep learning era \cite{netvlad}, VPR methodologies have remained closely aligned with those in Image Retrieval (IR) in computer vision and Loop Closure Detection (LCD) in robotics. The pipeline of recent methods can be summarized as follows:

\begin{figure}
\centering
\includegraphics[width=\linewidth]{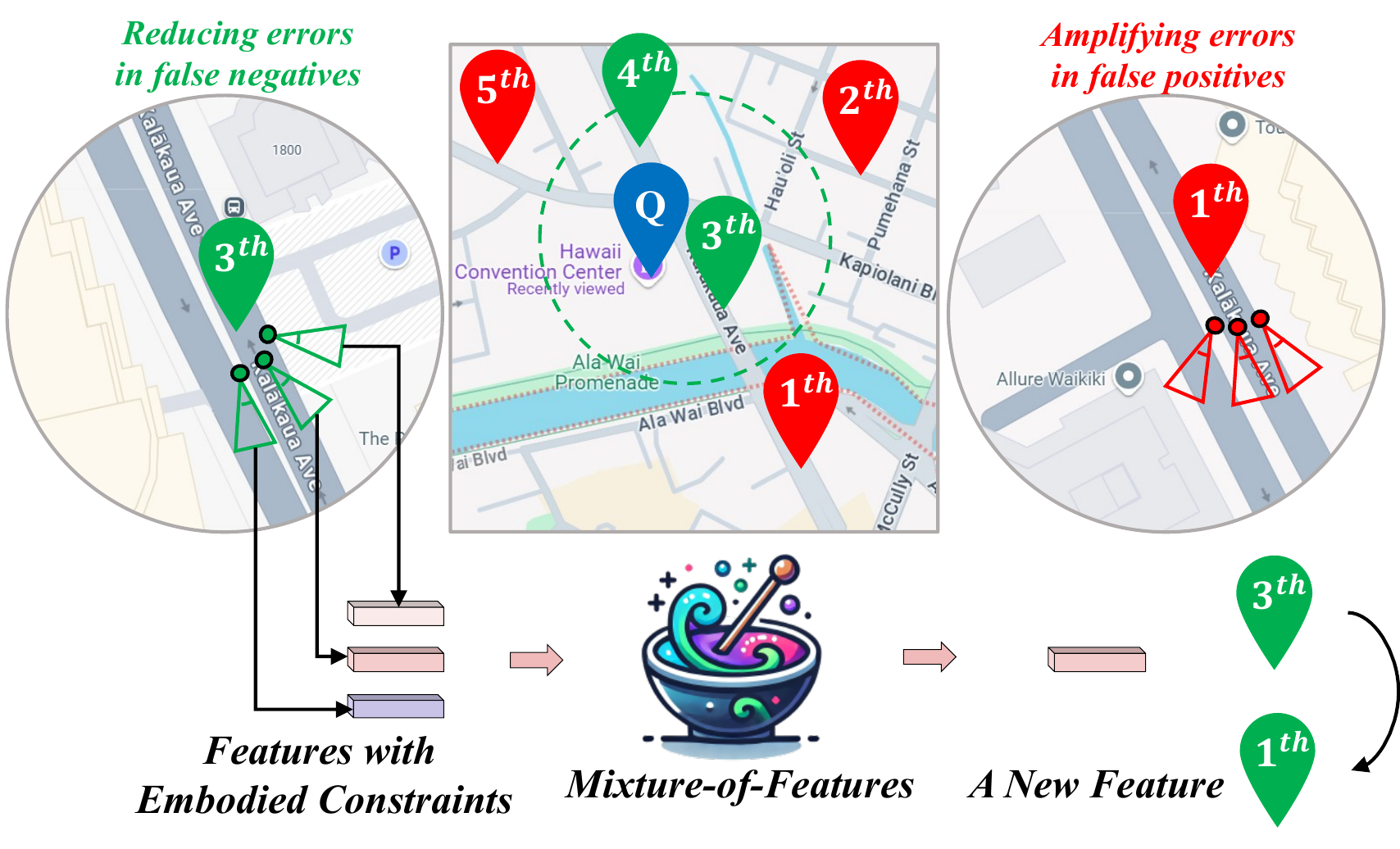}
\caption{\textbf{Mixture-of-Features with Embodied Constraints.} We illustrate an example of how \textit{EmbodiedPlace} works. In the middle subfigure, the candidate rankings do not correspond correctly to the geographic locations. By mixing features that satisfy embodied constraints, EmbodiedPlace refines the feature representation by reducing false-negative errors and converting them into true positives, while amplifying false-positive errors to suppress them correctly. As a result, the candidate rankings are adjusted to better align with the actual geographic locations.}
\label{fig:demo}
\end{figure}

\begin{enumerate}
\item \textbf{(Mandatory)} A neural network is trained on a large-scale image dataset using metric \cite{msloss} or classification loss functions \cite{cosface}, mapping images into a high-dimensional embedding space. During inference, the trained network extracts global features for both query and database images, followed by a KNN search in the feature database to retrieve the $K$ most similar results.
\item \textbf{(Optional)} A separate local feature extraction branch is trained, and the extracted local features are used for geometric verification or re-ranking of the $K$ candidate results during inference.
\end{enumerate}

The hierarchical relationship among these three tasks can be represented as: $\text{LCD} \subset \text{VPR} \subset \text{IR}$. To better understand the subtle and complex differences among these tasks, we analyze them from \textit{an application-based perspective}. While IR is a retrieval task for arbitrary images, both VPR and LCD focus on scene images. Additionally, as a module of SLAM \cite{mur2015orb}, LCD inherently adheres to a continuous motion model, making it natural for robotics researchers to incorporate spatiotemporal verification \cite{mk_lcd}. VPR is often applied as a sub-module of visual localization (VL) \cite{hloc} or an LCD module \cite{mk_lcd}. However, computer vision researchers often focus more on feature extraction, sometimes overlooking the potential of leveraging embodied information.

Another argument we present is that designing a dedicated local feature extraction branch for VPR is \textit{impractical}. Local feature extraction and matching algorithms (e.g., SuperPoint \cite{superpoint}, SuperGlue \cite{superglue}) are widely adopted in VL systems \cite{hloc}. These algorithms are likely to outperform local features specifically designed for VPR, as the latter are typically not evaluated on standard benchmarks such as HPatch \cite{balntas2017hpatches}. Furthermore, employing these established algorithms for re-ranking and verification in VPR is both feasible and efficient. Maintaining two distinct local feature extraction pipelines within a single system would introduce significant storage and computational overhead.

Although some methods \cite{superglue} have been proposed for verification or re-ranking without relying on local features, they often increase computational complexity and fail to leverage embodied constraints. In this work, we propose a simple yet carefully designed re-ranking method that adapts to various embodied constraints while operating solely on global features via a Mixture-of-Features (MoF) mechanism, as illustrated in \cref{fig:demo}.

\textbf{Our contributions} are summarized as follows:

\begin{enumerate}
\item We analyze the feasibility of embodied constraints in practical VPR applications and categorize their forms in existing public datasets.
\item We propose a re-ranking method for VPR that operates exclusively on global features under embodied constraints, leveraging feature mixtures from multiple images with data associations.
\item We explore a weight-computation strategy for MoF using a multi-metric learning approach.
\item Experimental results show that our method improves state-of-the-art (SOTA) performance on public datasets with minimal computational overhead.
\end{enumerate}

\section{Related Work}

\begin{table*}[ht]
\centering
\small
\begin{adjustbox}{width=\linewidth}
\begin{tabular}{@{}l|ccccccccc}
\toprule
Method & QE \cite{Chum_Philbin_Sivic_Isard_Zisserman_2007} & DBA \cite{Arandjelovic_Zisserman_2012} & SuperGlobal \cite{shao2023global} & GV \cite{patch_netvlad} & SV \cite{mk_lcd}  & RRT \cite{reranking}  & CV-Net \cite{lee2022correlation} & \textbf{EmbodiedPlace} \\
\hline
Global features & \checkmark & \checkmark & \checkmark & $\times$ & $\times$ &\checkmark & \checkmark & \checkmark \\ 
Local features & $\times$ & $\times$ & $\times$ & \checkmark & \checkmark & \checkmark & \checkmark &  $\times$ \\ 
Refine targets & Q & D & Q\&D & - & - & - & - & Subsets fo D\\
Embodied info. & $\times$ & $\times$ & $\times$ & $\times$ & Sequence & $\times$ & $\times$ & General  \\
Additional storage / time & Low & High & Low & High & High & High & High & Low \\
\bottomrule
\end{tabular}
\end{adjustbox}
\caption{\textbf{Comparison of Re-ranking Methods.} We present a comparison of representative re-ranking methods based on their characteristics. The table evaluates whether the refined features are global or local, whether refinement applies to query (Q) features or database (D) features, the type of embodied information utilized, and the relative computational/storage overhead of each method.}
\label{tab:comparison_methods}
\end{table*}

\subsection{One-stage Methods}
The evolution of research in VPR, IR, and LCD has followed a nearly parallel trajectory. Early approaches primarily relied on hand-crafted features, including directly extracted global features (e.g., GIST \citep{seqslam}) and features aggregated from local descriptors. Common aggregation techniques included Bag of Words (BoW) \citep{bow}, Fisher Vector (FV) \citep{fisher}, and Vector of Locally Aggregated Descriptors (VLAD) \citep{vlad}, often paired with local feature extraction algorithms such as SIFT \citep{sift}, SURF \citep{surf}, and ORB \citep{orb}.

The advent of deep learning has led to the widespread adoption of learned features, which have largely supplanted hand-crafted methods. \textit{As this paper does not focus on one-stage methods, we highlight only a few representative works.} Arandjelovic et al. \cite{netvlad} proposed a differentiable local residual aggregation layer and fine-tuned models on the Pittsburgh dataset using a triplet loss function. Radenovic et al. \cite{gem} introduced a generalized mean pooling (GeM) layer and fine-tuned models on the Retrieval-SFM dataset. More recently, BoQ and SALAD were proposed, both employing multi-similarity loss functions \cite{msloss} on the GSV-Cities dataset \cite{mixvpr}. BoQ \cite{boq} integrates an attention-based feature extractor with learnable query parameters, whereas SALAD \cite{salad} redefines the soft assignment of local features in NetVLAD as an optimal transport problem.

A notable method, CricaVPR \cite{crica}, introduced cross-image fusion for feature extraction and shares some similarities with our work. However, it presents two significant issues: (1) The joint features of $M$ images in a batch implicitly incorporate embodied information without explicitly analyzing its forms, making the value of $M$ and dataset variations substantially impact performance. (2) CricaVPR applies cross-image fusion to query features, which \textit{conflicts with the problem definition of VPR} and introduces potential test-time information leakage. Moreover, our method is implemented via a lightweight matrix plugin, distinguishing it from their attention modules, which are coupled within the network.

\subsection{Two-stage Methods}

Two-stage VPR methods first retrieve $K$ candidate images using global features and then re-rank them through local feature matching. However, in IR, some re-ranking approaches do not rely on local features.

\noindent \textbf{Visual Place Recognition:} Geometry Verification (GV) in VPR was traditionally performed using RANSAC, which validates candidate matches by analyzing local feature correspondences for spatial consistency. Patch-NetVLAD \cite{patch_netvlad} innovates by aggregating patch-level features derived from NetVLAD residuals, which enable scale-invariant matching. Recent works introduce transformer-based approaches that redefine GV. For example, R2Former \cite{r2} incorporates local feature correlations, attention values, and positional information into a unified retrieval and re-ranking pipeline. TransVPR \cite{transvpr} leverages multi-level attention aggregation to extract task-relevant global features and uses filtered transformer tokens for spatial matching. EffoVPR \cite{effovpr} show that features extracted from self-attention layers can act as a powerful re-ranker for VPR, even in a zero-shot setting. However, the coexistence of dedicated VPR local features and general-purpose local features in a single system leads to inefficiency, highlighting a disconnect between these studies and practical applications.

\noindent \textbf{Image Retrieval:} We focus on less common global feature-based re-ranking approaches. Query Expansion (QE) \cite{Chum_Philbin_Sivic_Isard_Zisserman_2007} enhances the representation of a query image by expanding it, thereby improving retrieval quality. Database-side Augmentation (DBA) \cite{Arandjelovic_Zisserman_2012} applies QE to each database image to achieve similar improvements. The Re-ranking Transformer (RRT) \cite{reranking, zhang2023etr} combines global and local features to re-rank candidates in a supervised manner, providing a more efficient alternative to traditional geometric verification methods. The Correlation Verification Network (CV-Net) \cite{lee2022correlation} employs stacked 4D convolutional layers to compress dense feature correlations, learning diverse geometric matching patterns. SuperGlobal \cite{shao2023global} refines global features by considering only the top-ranked images for partial re-weighting of features. We provide more detailed comparisons in \cref{tab:comparison_methods} and the supplementary materials.

\noindent \textbf{Loop Closure Detection:} In robotics, LCD methods often utilize the temporal and spatial continuity inherent in image sequences. These approaches frequently match local features across $N$ consecutive frames, with methods such as FILD \cite{an2019fast} and MK-LCD \cite{mk_lcd} exemplifying this strategy. In practice, embodied constraints tend to be more diverse and complex. For example, Google Street View collection services may revisit the same GPS location multiple times a year, often employing panoramic cameras.

\subsection{Others}

MoF is a simple yet novel technique that, although superficially resembling Mixture-of-Experts (MoE) \cite{riquelme2021scaling}, Mixup \cite{zhang2017mixup}, and feature fusion techniques \cite{wu2023asymmetric}, exhibits \textit{fundamental differences}. (1) MoE decomposes internal modules, whereas MoF blends external features. (2) Mixup is a data augmentation technique that operates on two input images during training, whereas MoF is a feature correction technique applied to multiple input features during both training and inference. (3) Feature fusion techniques are typically tightly integrated within neural network, where different features generated from a single input are fused, and features change during backpropagation. In contrast, MoF operates differently: multiple input features remain unchanged and are combined to generate a new feature representation.

\section{The Embodied Constraints in VPR}
\label{EC_VPR}

\begin{figure*}
\centering
\includegraphics[width=\textwidth]{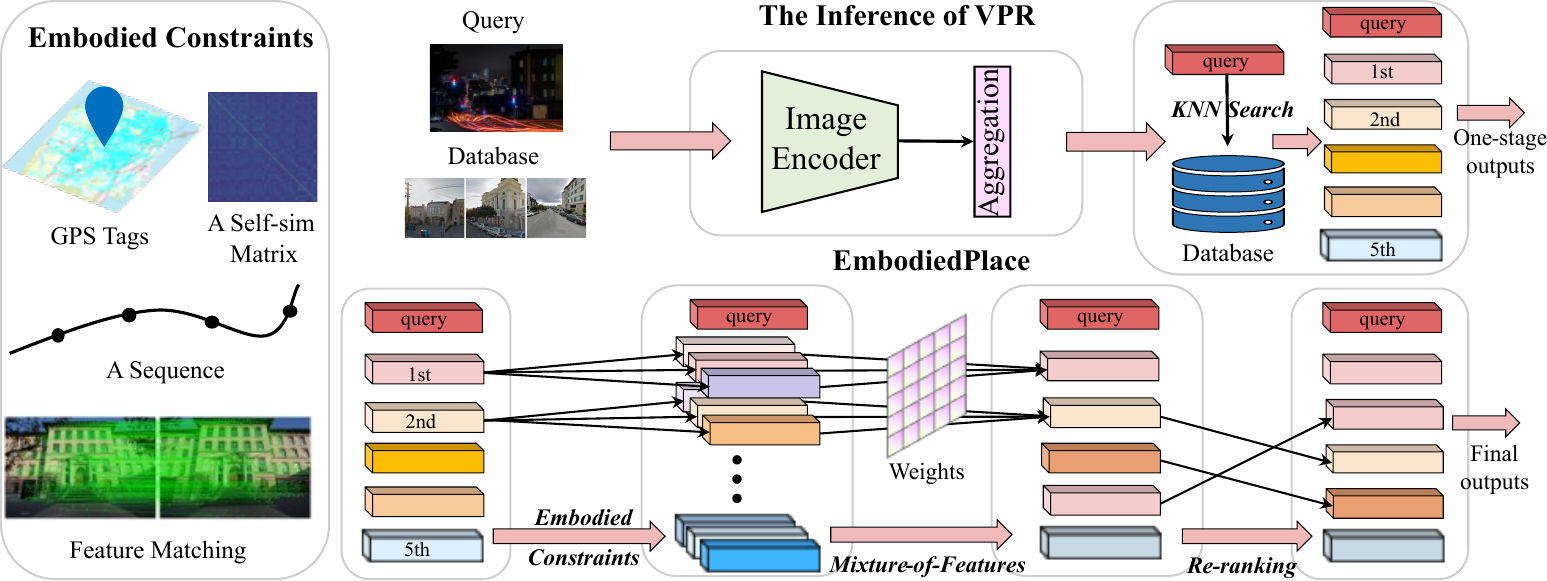}
\caption{\textbf{Illustrative Diagram of the Methodology.} This figure includes three components: the four forms of embodied constraints, the general VPR inference pipeline, and the inference pipeline of EmbodiedPlace.} 
\label{fig:framework}
\end{figure*}

Given a query image $I_\text{q}$, the task is to retrieve $K$ similar images from a large-scale image database $ \mathcal{I} = \{I_i\}_{i = 1}^{N} $. In addition to the above problem definition, the VPR task follows a standard protocol \cite{netvlad} in which each image $I_i \in \mathcal{I}$ is associated with a GPS tag $g_i \in \mathbb{R}^2$, representing its geographical location. A database image $I_i$ is considered a true positive for a query $I_\text{q}$ if its GPS tag $g_i$ meets a spatial proximity constraint relative to the ground-truth location $g_\text{q}$ of the query. For instance:
\begin{equation}
S(I_i, I_\text{q}) = 
\begin{cases} 
1 & \text{if } \|g_i - g_\text{q}\| \leq \epsilon, \\
0 & \text{otherwise},
\end{cases}
\end{equation}
where $\epsilon > 0$ is a predefined threshold, typically set to 25 meters.

\noindent \textbf{GPS tags} are referred to as an embodied constraint beyond visual information, or using classical terminology, as \textit{data associations}. Specifically, the database images are cropped from panoramic images, which constitute another embodied constraint. We argue that such embodied constraints within the database also exist in real-world applications, and \textit{GPS-denial conditions correspond to target query images}. Thus, we can introduce embodied constraints to enhance VPR performance during inference. Beyond the aforementioned GPS tags, other embodied constraints in VPR include:

\noindent \textbf{Timestamps of Sequences.} A robot's motion is continuous, and accordingly, the video captured by a camera is sequential. When the camera moves at a constant speed and displacement occurs along the main optical axis, timestamps can serve as a measure of image similarity. Considering nearly uniform motion as a simple example, we define a fixed similarity interval $[-t, t]$ and a dissimilarity interval $(-\infty, -t-t_m] \cup [t+t_m, \infty)$, where $t_m$ helps prevent confusion between similar and dissimilar images \cite{mk_lcd}. Formally, the similarity measure $S(I_i, I_j)$ based on timestamps $t_i$ and $t_j$ is defined as:
\begin{equation}
S(I_i, I_j) = 
\begin{cases} 
1 & \text{if } |t_i - t_j| \leq t, \\
0 & \text{if } |t_i - t_j| > t + t_m,
\end{cases}
\end{equation}

\noindent \textbf{Feature Matching.} In real-world applications, VPR serves as the first stage of a hierarchical visual localization system. Such systems typically include a pre-built point cloud map constructed using Structure-from-Motion (SfM) and known frame correspondences. These correspondences serve as an approximate form of embodied constraints due to their inherent imprecision. Although not entirely accurate, they provide valuable guidance for constraining potential pairs and improving retrieval. To quantify the quality of matches, we introduce the inlier-controlled similarity label $S$, defined as:
\begin{equation}
S(I_i, I_j) = 
\begin{cases} 
1 & \text{if } \frac{N_\text{inliers}}{N_\text{total}}  > \sigma, \\
0 & \text{otherwise},
\end{cases}
\end{equation}
where $N_\text{inliers}$ is the number of geometrically consistent matches and $N_\text{total}$ is the total number of matches.

\noindent \textbf{Self-similarity Matrix.} A self-similarity matrix $M$ captures pairwise feature similarity among database images. Formally, for a database of 
$N$ images, the matrix $M \in \mathbb{R}^{N \times N}$  is defined as: 
\begin{equation}
M_{i, j} = \text{sim}(f_i, f_j),
\end{equation}
where $\text{sim}(\cdot, \cdot)$ is a similarity function (e.g. Cosine similarity or Euclidean distance). This matrix can be leveraged as a pseudo-embodied constraint, applying thresholding to refine retrieval or re-ranking:
\begin{equation}
S(I_i, I_j) = 
\begin{cases} 
1 & \text{if } 1 > M_{i, j} > \delta, \\
0 & \text{otherwise},
\end{cases}
\end{equation}
where $\delta$ is a predefined similarity threshold. Such matrices are useful for identifying clusters or sequences of similar images, especially when combined with other embodied constraints, such as timestamps.

\section{EmbodiedPlace}
\label{sec:EP}

\subsection{Global Feature Extraction and Retrieval}

Global feature extraction and retrieval exist in all modern VPR methods. Meanwhile, the proposed method introduces an optional module within a general VPR framework rather than improving any single module. Consequently, the proposed method can theoretically be integrated with any model without modifying its parameters. Let the extracted query feature be represented as $f_\text{q}$, and the database features as $ \mathcal{F} = \{f_i\}_{i = 1}^{N} $. The feature dimension is $D$, and the resulting database has a dimension of $N \times D$. We use FAISS (Facebook AI Search System) \cite{douze2024faiss}, a vector-based data storage system that utilizes flat indexes and GPU parallel computation techniques.

Given a query feature $f_\text{q}$, a candidate set of $K$ features can be retrieved from the database. This candidate set has dimensions of $K \times D$, where $K$ is a constant and typically less than 1000. Formally, the set of candidates $C$ is defined as: $\mathcal{C} = \{f_{\text{c}_1}, f_{\text{c}_2},..., f_{\text{c}_k}\} \subset \mathcal{F}$.

\subsection{Selecting Neighbors of Candidates}

Combining the representations of similar images with the representation of the original image to form a refined representation is an existing technique. Specifically, SuperGlobal performs a secondary query to refine the top-$K$ candidate features. For each candidate feature $f_{\text{c}_i} \in \mathcal{C}$, there is a set of $L$ neighboring features $\mathcal{N}_i = \{f_{\text{n}_1}, f_{\text{n}_2},..., f_{\text{n}_L}\} \subset \mathcal{F}$, where $L$ is a constant and typically less than 10. These neighboring features are then used to adjust the $K$ candidate features, enabling feature re-ranking. However, as acknowledged in their work, \textit{false positives in the retrieved neighbors may harm the expanded representation}.

Since the query features in the secondary query are derived from the first query’s candidate features, this process is inherently self-contained \footnote{Unaffected by the query features used in the initial retrieval.}. Instead of conducting a secondary KNN search, we introduce embodied constraints, as described in \cref{EC_VPR}, to organize the selection of neighboring features. If the number of available neighboring features does not meet the required $L$, we duplicate $f_{\text{n}_1}$ to maintain a consistent list length.

\subsection{Mixture-of-Features}

Given a query feature $f_\text{q}$, a candidate feature $f_{\text{c}_i} \in \mathcal{C}$, and a set of $L$ neighboring features $\mathcal{N}_i$, the goal of MoF is to obtain the refined candidate feature $f'_{\text{c}_i}$. The MoF process can be described as follows:
\begin{equation}
f'_{\text{c}_i} = \sum_{j=1}^L w_j \cdot f_{\text{n}_j},
\end{equation}
where $f_{\text{n}_1}$ is identical to $f_{\text{c}_i}$ (i.e., the first neighboring feature). In previous studies, the weights $w_j$ were determined either through manually tuned parameters \cite{Chum_Philbin_Sivic_Isard_Zisserman_2007} or simpler semi-adaptive methods \cite{shao2023global}. However, these approaches often fail to achieve competitive results, particularly due to the saturation effect observed in recent methods. To address this limitation, we explore novel and effective methods for computing the MoF weights.

\begin{figure}
\centering
\includegraphics[width=\linewidth]{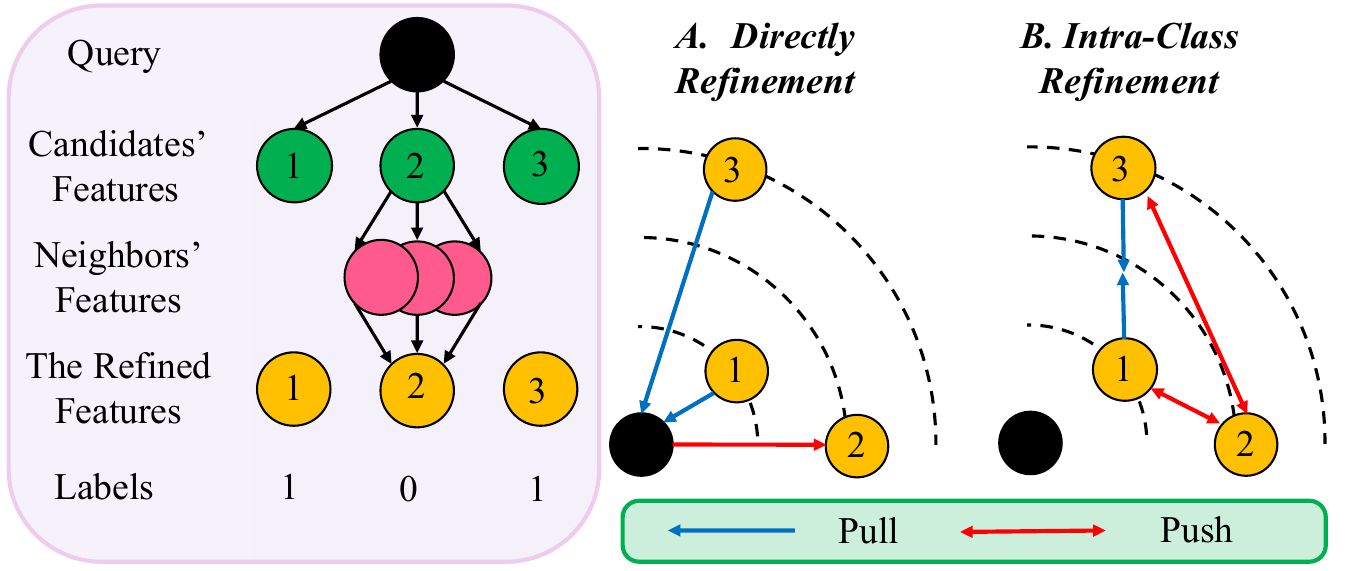}
\caption{\textbf{Losses for Feature Refinement.} This figure illustrates two metric loss functions designed for feature refinement. The first aims to minimize the distance to the query, while the second focuses on minimizing the distance between positive samples of the same class.}
\label{fig:loss}
\end{figure}

\noindent \textbf{Learning weights optimized by metric losses.} A learnable parameter matrix $W \in \mathbb{R}^{N \times D}$ is introduced as the weight model. This matrix is optimized using metric loss functions. The basic form of the metric loss is:
\begin{equation}
\mathcal{L} = \max(0, d(a, p) - d(a, n) + \alpha),
\end{equation}
where $a$ is the anchor sample, $p$ is the positive sample, $n$ is the negative sample, and $\alpha$ is a margin hyperparameter. We further decompose this loss into two sub-loss functions:

\textbf{1) The direct refinement loss} aims to minimize the Euclidean distance between the refined candidate feature $f'_{\text{c}_i}$ and the query feature $f_q$ for samples with a label of 1, while maximizing the distance for samples with a label of 0. The loss function is defined as:
\begin{equation}
\begin{aligned}
\mathcal{L}_\text{Direct} = \sum_{i=1}^N & \mathbb{I}(y_i=1) \cdot d(f'_{\text{c}i}, f_q) - \\ & \mathbb{I}(y_i=0) \cdot d(f'_{\text{c}_i}, f_q),
\end{aligned}
\end{equation}
where $\mathbb{I}(\cdot)$ denotes the indicator function.

\textbf{2) The intra-class refinement loss}  aims to minimize the Euclidean distance between refined candidate features $f'_{\text{c}_i}$ and $f'_{\text{c}_j}$ when both have label 1, and maximize the Euclidean distance between $f'_{\text{c}_i}$ and $f'_{\text{c}_k}$ when $f'_{\text{c}_i}$ has label 1 and $f'_{\text{c}_k}$ has label 0. The loss is formulated as:

\begin{equation}
\begin{aligned}
\mathcal{L}_\text{Intra} = \sum_{i \neq j}^N  & \mathbb{I}(y_i=1, y_j=1) \cdot d(f'_{\text{c}_i}, f'_{\text{c}_j}) - \\ & \mathbb{I}(y_i=1, y_k=0) \cdot d(f'_{\text{c}_i}, f'_{\text{c}_k}).
\end{aligned}
\end{equation}

The final loss function is a weighted combination of the sub-losses:
\begin{equation}
\mathcal{L}_{\text{total}} = \lambda_1 \mathcal{L}_\text{Direct} + \lambda_2 \mathcal{L}_\text{Intra},
\end{equation}
where $\lambda_1$ and $\lambda_2$ are hyperparameters controlling the contribution of each sub-loss.

\noindent \textbf{Other Methods.} We also explored several other promising methods, including (1) implementing MoF using cross-attention mechanisms \cite{attention}, (2) directly regressing rankings instead of adjusting features, and (3) using full-adaptive weights based on query similarity. However, these methods did not perform better than the learning-based MoF and are discussed in the supplementary materials.

\subsection{Reranking based on Global Features}

During the inference phase, we compute the Euclidean distance between the query feature $f_\text{q}$ and the refined candidate features $\mathcal{C'} = \{f'_{\text{c}_1}, f'_{\text{c}_2},..., f'_{\text{c}_k}\} \subset \mathcal{F'}$, and then re-rank the candidates accordingly.
\section{Experiments}

In this section, we present a comprehensive set of experiments to rigorously evaluate the effectiveness of the proposed method. First, we introduce the implementation details and evaluation protocol. Next, we provide a fine-grained comparison of EmbodiedPlace with SOTA methods, covering three categories: retrieval-based methods, global feature reranking methods, and local feature reranking methods. Finally, we discuss ablation studies and parameter sensitivity analyses to gain deeper insights into the contributions of our approach.

\subsection{Experimental Setup}

\noindent \textbf{Implementation Details.} Our proposed method can be easily integrated into any VPR model in a plug-and-play manner. Inspired by recent VPR approaches based on DINOv2 \cite{dinov2}, we implemented DINOv2-GeM (see the supplementary material for details) and applied EmbodiedPlace to it, as it achieves results comparable to recent works while using only ultra-low-dimensional features. To further demonstrate its \textit{generalizability}, we also conducted experiments by integrating it with BoQ \cite{boq}. The baseline model and the proposed modules in this paper are implemented in PyTorch, with both training and inference conducted on a single NVIDIA RTX 3090 GPU paired with an AMD EPYC-9754 CPU. Notably, our method is computationally lightweight, and its execution requirements are expected to be significantly lower than those of our experimental setup.

\noindent \textbf{Training Configurations.} We train EmbodiedPlace using the Adam optimizer with a learning rate of 0.003 and a batch size of 64. Training stops when the R@1 on the validation set does not improve for three consecutive epochs. Following previous work \cite{patch_netvlad, transvpr, selavpr, r2}, we train our model on the MSLS and Pitts-30k training sets.

\subsection{Evaluation Benchmarks}

\begin{table}
\begin{center}
\caption{\textbf{Overview of test sets.} We can see huge variations in size and types of embodied constraints across the datasets.}
\begin{adjustbox}{width=\linewidth}
\centering
\begin{tabular}{lcccccccccc}
\toprule
\multirow{2}{*}{Dataset} & \multirow{2}{*}{Description} & Embodied & \multicolumn{2}{c}{Number}  \\
& & info. &Q & D \\
\midrule
Pitts-30k / 250k & urban, panorama & GPS Tags & 7k / 8k & 10k / 83k  \\
MSLS &urban, long-term & Self-sim Matrix & 740 & 18871  \\
Nordland &suburban & Sequence & 27592 & 27592  \\
Aachen v1.1 & scenic spot & Feature matching & 1015 & 6697  \\
\bottomrule
\end{tabular}
\end{adjustbox}
\end{center}
\label{tab:comparison_test_sets}
\vspace{-0.6cm}
\end{table}

\noindent \textbf{Test sets.} We evaluate our method on several widely used VPR benchmark datasets, including Pitts-30k / 250k, MSLS, Nordland, and the Aachen v1.1 visual localization dataset, as summarized in \cref{tab:comparison_test_sets}.

\begin{itemize}
\item The Pittsburgh dataset \cite{netvlad} contains 30k / 250k reference images and 24k query images in its training, validation, and test sets, exhibiting significant viewpoint variations.
\item The Mapillary Street-Level Sequences (MSLS) dataset \cite{msls} consists of 1.6 million images captured in urban, suburban, and natural environments over seven years, introducing long-term appearance variations.
\item The Nordland dataset \cite{nordland} is characterized by seasonal variations, where sequences of images are taken along the same railway route in different seasons, testing robustness to environmental changes.
\item To validate feature matching as an embodied constraint, we evaluate our method on the Aachen v1.1 \cite{vl_benchmarking}, a visual localization dataset. The database images were taken during the daytime over two years using handheld cameras, while query images include both daytime and nighttime.
\end{itemize}
 
\noindent \textbf{Evaluation metrics.} Following standard evaluation protocols \citep{netvlad, dvg}, we measure Recall@$K$, defined as the percentage of query images for which at least one of the top-$K$ retrieved reference images lies within a predefined threshold distance. For datasets with GPS labels, we use a threshold of 25 meters. Specifically, for Aachen v1.1, we follow the evaluation setup of previous works \citep{hloc}.

\subsection{Comparison with Retrieval-based Methods}

\begin{table*}
\caption{\textbf{Comparison to SOTA retrieval-based methods on benchmark datasets.} $\dag$ The results reported by CricaVPR use multiple (16) query images, so we additionally report the results of a single query image. $^\ddag$ We reimplemented GeM based on DINOv2, see the supplementary material for details.}
\small
\centering
\begin{adjustbox}{width=\textwidth}
\begin{tabular}{@{}lcccccccccccccc}
\toprule
\multirow{2}{*}{Method}  & \multirow{2}{*}{Vene} &  Feat.  & \multicolumn{3}{c}{Pitts-30k-test}  & \multicolumn{3}{c}{MSLS-val} & \multicolumn{3}{c}{Pitts-250k-test} & \multicolumn{3}{c}{Nordland} \\
& & dim. & R@1 & R@5 & R@10 & R@1 & R@5 & R@10 & R@1 & R@5 & R@10 & R@1 & R@5 & R@10 \\
\midrule
\multicolumn{15}{l}{\textbf{Global feature retrieval}} \\
NetVLAD & \textit{CVPR'16} &32768  & 81.9 &91.2 &93.7  & 53.1 &66.5 &71.1 & 90.5 & 96.2 & 97.4 & 13.1 & - & - \\
SFRS & \textit{ECCV'20} & 4096 & 89.4 & 94.7 & 95.9 & 69.2 & 80.3 & 83.1 & 90.1 & 95.8 & 97.0 & 16.0 &24.1 &28.7  \\
CosPlace & \textit{CVPR'22} & 2048 & 90.9 & 95.7 & 96.7 &  87.2 & 94.1 & 94.9 & 92.3 & 97.4 & 98.4 & 58.5 &73.7 &79.4\\
MixVPR & \textit{WACV'22} & 4096 & 91.5 & 95.5 & 96.3 & 88.0 & 92.7 & 94.6 & 94.2 & 98.2 & 98.9 & 76.2 & 86.9 & 90.3\\
EigenPlaces & \textit{ICCV'23} &2048 & 92.5 & 96.8 & 97.6 & 89.1 & 93.8 & 95.0  & 94.1 & 97.9 & 98.7 & 71.2 & 83.8 & 88.1  \\
CricaVPR-16 & \textit{CVPR'24} & 4096 & 94.9$^\dag$ & 97.3 & 98.2  &  90.0 & 95.4 & 96.4 & - & - & -& 90.7 & 96.3 & 97.6 \\
CricaVPR-1 & \textit{CVPR'24} & 4096 & 91.6 & 95.7 & 96.9   & 88.5 & 95.1 & 95.7 & - & - & -& - & - & - \\
SALAD & \textit{CVPR'24} & 8448 & 92.4 & 96.3 & 97.4 &  92.2 & 96.2 & 97.0 & 95.1 & 98.5 & 99.1 & 89.7 & 95.5 & 97.0 \\
BoQ & \textit{CVPR'24} & 12288 & 93.7 & 97.1 & 97.9  & 93.8 & 96.8 & 97.0 & 96.6 & 99.1 & 99.5 & 90.6 & 96.0 & 97.5 \\
GeM$^\ddag$  & \textit{PAMI'18} & \textbf{768} & 91.8  & 96.5  & 97.4& 91.8 & 96.8 & 97.3 & 94.7 & 98.8  & 99.4 & 82.7 &92.5 & 95.0 \\
\midrule
\multicolumn{15}{l}{\textbf{Global feature retrieval + Global feature reranking}} \\
GeM + EmbodiedPlace & \textit{Ours} & \textbf{768} & 92.7 (\textbf{+0.9}) & 96.5  & 97.4  & 92.8 (\textbf{+1.0}) & 96.8 & 97.3 & 95.6 (\textbf{+0.9}) & 98.8  & 99.4& 84.2 (\textbf{+1.5}) & 92.5 & 95.0 \\
\bottomrule
\end{tabular}
\end{adjustbox}
\vspace{-0.3cm}
\label{tab:comparison_sota}
\end{table*}

As shown in \cref{tab:comparison_sota}, we compare EmbodiedPlace with SOTA retrieval-based methods, including NetVLAD \cite{netvlad}, SFRS \cite{sfrs}, CosPlace \cite{cosface}, MixVPR \cite{mixvpr}, EigenPlaces \cite{eigenplaces}, CricaVPR \cite{crica}, SALAD \cite{salad}, BoQ \cite{boq}, and GeM \cite{gem}. One of the key strengths of GeM is its ability to achieve results comparable to those of recent SOTA methods while maintaining an ultra-low feature dimensionality of 768. EmbodiedPlace builds on this strength and further improves performance across all test datasets. Although EmbodiedPlace slightly lags behind BoQ (12,288-dim), it outperforms other high-dimensional feature-based methods such as SALAD.

On the Pitts-30k-test dataset, our method improves the R@1 performance by 0.9\%, reaching 92.7\%, compared to GeM's 91.8\%. Notably, CricaVPR reports an R@1 of 94.9\% on Pitts-30k; however, this result is influenced by query leakage issues. By evaluating EmbodiedPlace on Nordland and MSLS, we demonstrate its strong cross-domain generalization, proving its effectiveness beyond a single type of embodied constraint. This indicates that our method is adaptable to various environmental conditions and dataset characteristics, reinforcing its robustness across different VPR scenarios. Furthermore, EmbodiedPlace performs exceptionally well on larger-scale datasets. On the Pitts-250k test, it achieves an R@1 score of 95.6\%, marking a 0.9\% improvement over GeM’s 94.7\%. This highlights the scalability and robustness of our reranking approach in large-scale retrieval tasks. Since most VPR methods are not evaluated on Aachen, we do not discuss it here in detail but instead demonstrate the feasibility of feature matching as an embodiment constraint, as shown in \cref{tab:aachen}.

\subsection{Comparison with Reranking-based Methods}

\begin{table}
\caption{\textbf{Comparison to global feature re-ranking methods on benchmark datasets.} In the first row, we report the R@1 and R@5 results for DINOv2-GeM. Time is calculated on the MSLS-Val dataset, which includes around 18,871 images.}
\small
\centering
\begin{adjustbox}{width=\linewidth}
\begin{tabular}{@{}lccccc}
\toprule
\multirow{2}{*}{Method} & Reranking & Pitts-30k & MSLS & Pitts-250k & \multirow{2}{*}{Nordland} \\
 & (ms) & test & val & test &  \\
\midrule
DINOv2-GeM & 0  & 91.8 / 96.5  & 91.8 / 96.8 & 94.7 / 98.8 & 82.7 / 92.5  \\
+ QE & 0.08 & 91.8 (+0.0) & 91.8 (+0.0) & 94.7 (+0.0) & 82.9 (+0.2) \\
+ DBA & 25.6 & 92.4 (+0.6) & 92.4 (+0.6) & 95.3 (+0.6) & 83.2 (+0.5) \\
+ SuperGlobal & 7.9 & 90.0 (-1.8) & 90.5 (-1.3) & 92.8 (-1.9) & 76.1 (-6.6) \\
+ EmbodiedPlace & 0.01 & 92.7 (\textbf{+0.9})  & 92.8 (\textbf{+1.0}) & 95.6 (\textbf{+0.9})& 84.2 (\textbf{+1.5})  \\
\bottomrule
\end{tabular}
\end{adjustbox}
\label{tab:comparison_global}
\vspace{-0.3cm}
\end{table}

\noindent \textbf{1) Global Feature Reranking.} To the best of our knowledge, this is the \textit{first} comparative study on global feature re-ranking for VPR in the past five years, as presented in \cref{tab:comparison_global}. We carefully reproduced and optimized for the best recall (though it may require significant computational overhead) for QE \cite{Chum_Philbin_Sivic_Isard_Zisserman_2007}, DBA \cite{Arandjelovic_Zisserman_2012}, and SuperGlobal \cite{shao2023global}, with further implementation details provided in the supplementary materials.

First, we evaluate the impact of QE, which does not enhance the retrieval performance of DINOv2-GeM. While QE was effective in earlier works, we hypothesize that its ineffectiveness here results from the strong performance of modern baselines, creating a saturation effect in which traditional refinements no longer yield noticeable gains.

Next, we assess DBA, which improves retrieval accuracy by refining global feature representations for all database images. Although DBA improves retrieval performance by +0.6\% R@1 across all datasets, its high memory and computational overhead make it impractical for large-scale VPR applications.

In contrast, SuperGlobal leads to a performance drop across all datasets. Before proposing EmbodiedPlace, we observed these experimental results which align with the SuperGlobal authors' statement that false positives in retrieved neighbors can negatively impact expanded representations. This issue appears to be exacerbated in large-scale VPR datasets. Notably, even with extensive parameter tuning, we were unable to achieve competitive results with SuperGlobal. (Please see the supplementary materials.)

We also performed similar experiments using DINOv2-BoQ, observing similar trends in the results. However, since BoQ already achieves strong performance, the recall improvement from EmbodiedPlace is relatively marginal. Additionally, due to the high dimensionality of BoQ features (12,288-dim), global feature refinement operations such as QE, DBA, and SuperGlobal scale linearly on the CPU, whereas the learning-based EmbodiedPlace remains efficient.

\begin{table}
\caption{\textbf{Comparison to global feature re-ranking methods on benchmark datasets.} In the first row, we report the R@1 and R@5 results for BoQ. The symbol "$\times$" indicates that our setup was unable to support testing this method on the respective dataset. }
\small
\centering
\begin{adjustbox}{width=\linewidth}
\begin{tabular}{@{}lccccc}
\toprule
\multirow{2}{*}{Method} & Reranking & Pitts-30k & MSLS & Pitts-250k & \multirow{2}{*}{Nordland} \\
 & (ms) & test & val & test &  \\
\midrule
DINOv2-BoQ & 0 & 93.7 / 97.1 & 93.8 / 96.8 & 96.6 / 99.1 & 90.6 / 96.0   \\
+ QE & 2.2 & 93.8 (+0.1) & 93.9 (\textbf{+0.1}) & 96.6 (+0.0) & 90.6 (+0.0)  \\
+ DBA & 705.59 & 94.1 (\textbf{+0.4}) & 92.3 (-1.6) & $\times$ & 90.0 (-0.6)  \\
+ SuperGlobal & $\times$ & $\times$ & $\times$ & $\times$ & $\times$  \\
+ EmbodiedPlace & 0.11 & 94.0 (+0.3) & 93.8 (+0.0) & 96.9 (\textbf{+0.3}) & 91.1 (\textbf{+0.5})   \\
\bottomrule
\end{tabular}
\end{adjustbox}
\label{tab:comparison_global_boq}
\vspace{-0.3cm}
\end{table}

\begin{table*}
\caption{\textbf{Comparison to local features reranking-based methods on benchmark datasets.} The storage consumption and latency values are taken from the paper \cite{r2}and the discussion of EffoVPR on OpenReview. Time and memory footprint are calculated on the MSLS-Val dataset, which includes around 18,871 images.}
\small
\centering
\begin{adjustbox}{width=\textwidth}
\begin{tabular}{@{}lcccccccccccccc}
\toprule
\multirow{2}{*}{Method} &  \multicolumn{2}{c}{Feat. dim.} & \multirow{2}{*}{Memory (GB)} & \multicolumn{2}{c}{Latency (ms)} & \multirow{2}{*}{Device} & \multicolumn{3}{c}{Pitts-30k-test}  & \multicolumn{3}{c}{MSLS-val}  \\
& Global & Local & & Retrieval &  Reranking&  & R@1 & R@5 & R@10 & R@1 & R@5 & R@10 \\
\midrule
\multicolumn{9}{l}{\textbf{Global feature retrieval + Local feature reranking}} \\
Patch-NV (\textit{CVPR'21}) & 4096 & 2826$\times$4096 & 908.3 & 0.19 & 8377.17  & A5000 & 88.7 & 94.5 & 95.9 & 79.5 & 86.2 & 87.7    \\
 TransVPR (\textit{CVPR'22}) & 256 & 1200$\times$256 & 22.72 & 0.07 & 1757.70 & A5000 &  89.0 &  94.9 &  96.2 & 86.8 & 91.2 & 92.4  \\
R2Former (\textit{CVPR'23}) & 256 & 500$\times$(128+3) & 4.79 & 0.07 & 202.37 & A5000 &91.1 &95.2 &96.3 &89.7 &95.0 &96.2 \\
SelaVPR (\textit{ICLR'24}) & 1024 & 61$\times$61$\times$ 128 & 3.5 & - & 85 & 3090&   92.8 & 96.8 & 97.7 & 90.8 & 96.4 & 97.2    \\
EffoVPR (\textit{ICLR'25}) & 1024 & 648$\times$1024 & 4.8 & - & 35 & A100 & 93.9 & 97.4 & 98.5 & 92.8 & 97.2 & 97.4  \\
\midrule
\multicolumn{9}{l}{\textbf{Global feature retrieval + Global feature reranking}} \\
GeM + EmbodiedPlace & \textbf{768} & \textbf{0} & \textbf{0.05} & 0.8 $^*$ & \textbf{0.01} & 3090 & 92.7 & 96.5  & 97.4  & \textbf{92.8} & 96.8 & 97.3 \\
\bottomrule
\end{tabular}
\end{adjustbox}
\label{tab:comparison_local}
\end{table*}

\noindent \textbf{2) Local Feature Reranking.} Table~\ref{tab:comparison_local} presents a comparison of the performance and computational efficiency of our method, EmbodiedPlace, against SOTA local feature re-ranking methods, including Patch-NV \cite{patch_netvlad}, TransVPR \cite{transvpr}, R2Former \cite{r2}, SelaVPR \cite{selavpr}, and EffoVPR \cite{effovpr}.

EmbodiedPlace demonstrates exceptional efficiency in terms of memory usage and latency. With a feature dimension of 768, it requires just 0.05 GB of memory and achieves an extremely low re-ranking latency of 0.01 ms. In stark contrast, other methods, such as Patch-NV and TransVPR, consume significantly more memory (e.g., 908.3 GB for Patch-NV) and have considerably higher latency (e.g., 8377.17 ms for Patch-NV). $^*$ Retrieval time is approximately proportional to the global feature dimension. The slightly longer retrieval time compared to R2Former maybe attributed to the relatively lower CPU performance of our setup. \footnote{We rented a card from a cloud server, so less CPU were allocated.} In theory, the retrieval time of EmbodiedPlace should fall between Patch-NV and R2Former.

Despite its low computational cost, EmbodiedPlace achieves competitive performance on both the Pitts-30k-test and MSLS-val datasets. For example, on the Pitts-30k test, our method achieves R@1 = 92.8\%. This outperforms several existing methods that involve more complex local feature re-ranking, such as Patch-NV (R@1 = 88.7\%) and TransVPR (R@1 = 89.0\%).

While methods like EffoVPR and SelaVPR achieve high retrieval accuracy (e.g., EffoVPR with R@1 = 93.9\% on Pitts-30k-test), they demand significantly greater memory and computational resources. EffoVPR, for instance, has a larger memory footprint (4.8 GB) and higher latency, even when using an A100 GPU.

\subsection{Ablation Study}

\begin{table}
\caption{\textbf{Performance comparison on Aachen-v1.1.}}
\centering
\small
\begin{adjustbox}{width=\linewidth}
\begin{tabular}{@{}lccccccc}
\toprule
Method  & \multicolumn{3}{c}{Day}  &  \multicolumn{3}{c}{Night}  \\
\midrule
GeM & \multicolumn{3}{c}{83.0 / 91.9 / 96.8}  & \multicolumn{3}{c}{65.4 / 86.4 / 96.9}  \\
GeM + EmbodiedPlace & \multicolumn{3}{c}{84.8 / 90.9 / 96.6}  & \multicolumn{3}{c}{67.5 / 84.3 / 96.9} \\
\bottomrule
\end{tabular}
\end{adjustbox}
\label{tab:aachen}
\end{table}

\begin{table}
\caption{\textbf{Ablation of the feature refinement losses.}}
\vspace{-0.1cm}
\centering
\small
\begin{adjustbox}{width=\linewidth}
\begin{tabular}{@{}cccccccc}
\toprule
Direct   & Intra-class  & Pitts-30k-test & MSLS-val  & Pitts-250k-test & Nordland \\
Refinement& Refinement& test & val & test &  \\\midrule

\checkmark & & \textbf{92.7} & 92.3 & \textbf{95.6} & \textbf{84.2} \\
 & \checkmark & 79.5 & 73.2 & 81.5 & 54.2 \\
\checkmark & \checkmark  & 91.9 & \textbf{92.8} & 94.7 & 81.6 \\
\bottomrule
\end{tabular}
\end{adjustbox}
\label{tab:ablation_loss}
\end{table}

\begin{table}
\caption{\textbf{Performance comparison with $K=5$.}}
\vspace{-0.1cm}
\centering
\begin{adjustbox}{width=\linewidth}
\label{tab:performance}
\begin{tabular}{clllll}
\toprule
Method & Pitts-30k-val & Pitts-30k-test & Pitts-250k-val & Pitts-250k-test \\
\midrule
Baseline & 93.9 & 91.8 & 93.5 & 94.7 \\
2 & 94.1\,(+0.2) & 91.8\,(+0.0) & 93.7\,(+0.2) & 94.7\,(+0.0) \\
4 & 94.4\,(+0.5) & 92.5\,(+0.7) & 94.0\,(+0.5) & 95.3\,(+0.6) \\
6 & \textbf{94.5\,(+0.6)} & 92.6\,(+0.8) & \textbf{94.1\,(+0.6)} & 95.5\,(+0.8) \\
8 & 94.4\,(+0.5) & \textbf{92.7\,(+0.9)} & 94.0\,(+0.5) & 95.6\,(+0.9) \\
10 & 94.4\,(+0.5) & 92.6\,(+0.8) & 94.0\,(+0.5) & 95.6\,(+0.9) \\
12 & 94.3\,(+0.4) & 92.6\,(+0.8) & 94.0\,(+0.5) & 95.6\,(+0.9) \\
16 & 94.3\,(+0.4) & 92.6\,(+0.8) & 94.0\,(+0.5) & \textbf{95.6\,(+0.9)} \\
\bottomrule
\end{tabular}
\end{adjustbox}
\label{tab:ablation_neighbors}
\vspace{-0.3cm}
\end{table}

\begin{table}
\caption{\textbf{Performance comparison with $L=8$.}}
\centering
\small
\begin{adjustbox}{width=\linewidth}
\begin{tabular}{clllll}
\toprule
Method & Pitts-30k-val & Pitts-30k-test & Pitts-250k-val & Pitts-250k-test \\
\midrule
Baseline & 93.9 & 91.8 & 93.5 & 94.7 \\
3 & 94.4\,(+0.5) & 92.5\,(+0.7) & 94.0\,(+0.5) & 95.4\,(+0.7) \\
5 & 94.5\,(\textbf{+0.6}) & 92.6\,(+0.8) & 94.1\,(\textbf{+0.6}) & 95.5\,(+0.8) \\
10 & 94.2\,(+0.3) & 92.7\,(\textbf{+0.9}) & 93.7\,(+0.2) & 95.6\,(\textbf{+0.9}) \\
30 & 93.4\,(-0.5) & 92.2\,(+0.4) & 93.0\,(-0.5) & 95.2\,(+0.5) \\
50 & 93.4\,(-0.5) & 92.1\,(+0.3) & 93.0\,(-0.5) & 95.2\,(+0.5) \\
\bottomrule
\end{tabular}
\end{adjustbox}
\label{tab:ablation_ranks}
\end{table}

As shown in \cref{tab:ablation_loss}, we evaluate the contribution of each term in the loss function. On the Pittsburgh dataset, the refinement loss alone achieved the best performance. However, on the MSLS dataset, combining both losses yielded better results. This difference is related to the strength of the embodied constraints. The GPS tags in Pittsburgh impose strong constraints, whereas the self-similarity matrix in MSLS provides a weaker constraint.

As shown in \cref{tab:ablation_neighbors} and \cref{tab:ablation_ranks}, we further examine the key parameters of EmbodiedPlace: the re-ranking size $K$ and the number of neighbors $L$. Due to the existing embodied constraints, increasing the number of neighbors does not produce negative effects but also does not lead to significant performance gains. However, expanding the re-ranking range can have negative effects, distinguishing it from local feature-based re-ranking. This is because the robustness of local features enables high-risk, high-reward re-ranking among the top-100 or even top-1000 retrieved candidates. In contrast, global feature re-ranking must be more conservative, typically restricted to the top-5 or top-10 candidates.

\section{Conclusion}

Building on insights from image retrieval and loop closure detection, we introduce a plug-and-play global feature re-ranking method for VPR. Specifically, we identify and categorize four types of embodied constraints commonly found in VPR datasets: GPS tags, sequential timestamps, feature matching-based pairings, and self-similarity matrix-based pairings. By selecting neighboring features that satisfy these embodied constraints and integrating them to refine candidate features, we introduce a novel Mixture-of-Features approach. The weighting of these mixed features is learned using a multi-metric loss function to optimize re-ranking effectiveness. Extensive experiments demonstrate that our method significantly improves the performance of existing VPR approaches while maintaining a low computational overhead.
{
    \small
    \bibliographystyle{ieeenat_fullname}
    \bibliography{main}
}
\clearpage

\newpage

\section{Supplementary Materials}

\subsection{Implementation Details of Our Baseline}

Inspired by the comparative experiments in SALAD \cite{salad}, we fine-tune DINOv2-GeM \cite{dinov2, gem} on the GSV-Cities \cite{mixvpr} dataset by using the MS-Loss function \cite{msloss}. The key difference is that we set the output dimension to the channel dimension of DINOv2 (768 dimensions) instead of 4096 or other values. The training batch size is 128, the optimizer is Adam, and the learning rate is set to $6 \times 10^{-5}$.

\subsection{Implementation Details of QE}

Query Expansion (QE) is a relatively underexplored technique in image retrieval that enhances query performance by refining the initial search results. The fundamental idea is to modify the original query by incorporating information from its nearest neighbors in the retrieval space. As shown in \cref{tab:qe_performance}, we conducted extensive parameter tuning experiments to compare and determine \textit{the optimal QE configuration}.

\noindent \textbf{Algorithm Description:}  
Given an initial query $q$, we retrieve its top-$k$ nearest neighbors from the database. The query representation is then updated by aggregating these neighbors' features, typically using an averaging function. The updated query is used to perform a second retrieval step, yielding improved results.

\begin{table}
\caption{\textbf{Performance comparison of QE on different datasets.}}
\centering
\small
\begin{adjustbox}{width=\linewidth}
\begin{tabular}{clllll}
\toprule
Method & Pitts-30k-test & MSLS-val & Pitts-250k-test & Nordland \\
\midrule
\multicolumn{5}{c}{\textbf{Fixed Parameters: $\beta = 0.9$}} \\
Baseline & 91.8 & 91.8 & 94.7 & 82.7 \\
5 & 91.84 & 91.76 & 94.66 & 82.86 \\
10 & 91.7 & 91.76 & 94.61 & 82.73 \\
25 & 91.59 & 91.62 & 94.49 & 82.44 \\
50 & 91.67 & 91.62 & 94.5 & 82.27 \\
100 & 91.7 & 91.76 & 94.5 & 82.22 \\
\midrule
\multicolumn{5}{c}{\textbf{Fixed Parameters: $\beta = 0.7$}} \\
Baseline & 91.8 & 91.8 & 94.7 & 82.7 \\
5 & 91.81 & 91.89 & 94.59 & 82.34 \\
10 & 91.51 & 91.49 & 94.47 & 81.8 \\
25 & 91.4 & 90.68 & 94.23 & 80.69 \\
50 & 91.3 & 90.81 & 94.1 & 80.02 \\
100 & 91.27 & 91.22 & 94.1 & 79.6 \\
\midrule
\multicolumn{5}{c}{\textbf{Fixed Parameters: $\beta = 0.5$}} \\
Baseline & 91.8 & 91.8 & 94.7 & 82.7 \\
5 & 90.99 & 91.89 & 93.85 & 80.91 \\
10 & 90.52 & 91.35 & 93.44 & 79.64 \\
25 & 90.05 & 89.46 & 92.89 & 76.87 \\
50 & 89.77 & 88.65 & 92.6 & 74.76 \\
100 & 89.83 & 89.32 & 92.9 & 73.29 \\
\midrule
\multicolumn{5}{c}{\textbf{Fixed Parameters: $\beta = 0.3$}} \\
Baseline & 91.8 & 91.8 & 94.7 & 82.7 \\
5 & 89.96 & 91.35 & 92.74 & 79.08 \\
10 & 88.63 & 89.73 & 91.7 & 75.92 \\
25 & 87.05 & 86.89 & 90.51 & 69.7 \\
50 & 86.59 & 85.95 & 89.8 & 64.28 \\
100 & 86.49 & 86.89 & 89.7 & 59.48 \\
\bottomrule
\end{tabular}
\end{adjustbox}
\label{tab:qe_performance}
\end{table}

\begin{algorithm}
\caption{Pseudocode of QE.}
\label{alg:qe}
\algcomment{\fontsize{7.2pt}{0em}\selectfont 
\texttt{retrieve}: function to find top-k nearest neighbors; \texttt{mean}: averaging function.
}
\definecolor{codeblue}{rgb}{0.25,0.5,0.5}
\lstset{
  backgroundcolor=\color{white},
  basicstyle=\fontsize{7.2pt}{7.2pt}\ttfamily\selectfont,
  columns=fullflexible,
  breaklines=true,
  captionpos=b,
  commentstyle=\fontsize{7.2pt}{7.2pt}\color{codeblue},
  keywordstyle=\fontsize{7.2pt}{7.2pt},
}
\begin{lstlisting}[language=python]
# q: query feature vector (1xC)
# D: database feature matrix (NxC)
# k: number of nearest neighbors
# beta: a weighting factor

# Step 1: Retrieve top-k nearest neighbors
N_q = retrieve(q, D, k)  # (kxC)

# Step 2: Compute expanded query representation
q_e = beta * q + (1 - beta) * mean(N_q, dim=0)# (1xC)

# Step 3: Perform retrieval again using q_exp
R_QE = retrieve(q_e, D, top_n)

return R_QE
\end{lstlisting}
\end{algorithm}

\subsection{Implementation Details of DBA}

Database-side Augmentation (DBA) is a retrieval enhancement technique that improves database representations before the query stage. Instead of modifying the query, DBA refines database embeddings by aggregating local neighborhood information. This helps in reducing noise and reinforcing the most relevant features. As shown in \cref{tab:dba_performance}, we conducted extensive parameter tuning experiments to compare and determine \textit{the optimal DBA configuration}.

\noindent \textbf{Algorithm Description:}  
Each database image representation is updated by incorporating feature information from its top-$k$ nearest neighbors. The motivation is to smooth the feature space by reinforcing similar embeddings, which improves retrieval robustness.
\begin{table}
\caption{\textbf{Performance comparison of DBA on different datasets.}}
\centering
\small
\begin{adjustbox}{width=\linewidth}
\begin{tabular}{clllll}
\toprule
Method & Pitts-30k-test & MSLS-val & Pitts-250k-test & Nordland \\
\midrule
Baseline & 91.8 & 91.8 & 94.7 & 82.7 \\
5 & 92.22 & 91.35 & 95.1 & 83.3 \\
10 & 92.37 & 91.35 & 95.3 & 83.2 \\
25 & 92.15 & 91.76 & 95.13 & 82.98 \\
50 & 92.18 & 92.03 & 95.1 & 83.1 \\
100 & 92.2 & 92.43 & 95.0 & 83.1 \\
\bottomrule
\end{tabular}
\end{adjustbox}
\label{tab:dba_performance}
\end{table}
\begin{algorithm}[t]
\caption{Pseudocode of DBA.}
\label{alg:dba}
\algcomment{\fontsize{7.2pt}{0em}\selectfont 
\texttt{retrieve}: function to find top-k nearest neighbors; \texttt{mean}: averaging function.
}
\definecolor{codeblue}{rgb}{0.25,0.5,0.5}
\lstset{
  backgroundcolor=\color{white},
  basicstyle=\fontsize{7.2pt}{7.2pt}\ttfamily\selectfont,
  columns=fullflexible,
  breaklines=true,
  captionpos=b,
  commentstyle=\fontsize{7.2pt}{7.2pt}\color{codeblue},
  keywordstyle=\fontsize{7.2pt}{7.2pt},
}
\begin{lstlisting}[language=python]
# D: database feature matrix (NxC)
# k: number of nearest neighbors

D_aug = zeros_like(D)  # initialize augmented database

# Compute augmented features for each database entry
for i in range(len(D)):
    N_i = retrieve(D[i], D, k)  # (kxC)
    D_aug[i] = mean(cat([D[i], N_i], dim=0), dim=0)  # (1xC)

# Replace database with augmented database
D = D_aug

return D
\end{lstlisting}
\end{algorithm}

\subsection{Explored Other Methods}

We also explored several \textit{promising} methods, which we share here \textit{for reference in future research}.

\begin{itemize}
    \item MoF using cross-attention mechanisms,
    \item Directly regressing rankings instead of adjusting features,
    \item Adaptive weights based on query similarity.
\end{itemize}

However, the first method performed even worse than the baseline in terms of recall, while the second method yielded only marginal improvements. The third method occasionally achieved promising results, but it appeared to require strong embodied constraints. Since the first two methods did not show significant improvements in our early experiments, we did not keep a record of their results. \textit{Here, we report only experiments on the last method}, as shown in \cref{tab:adaptive_embodiedplace}.

\noindent \textbf{MoF using Cross-Attention Mechanisms.}  
We applied a multi-head cross-attention mechanism to model interactions between neighboring features in the feature space. The idea was to refine candidate features by dynamically attending to their nearest neighbors. However, this approach did not yield improvements, likely because the popular attention module was too complex for refining global features.

\noindent \textbf{Directly Regressing Rankings.}  
Instead of refining feature embeddings, we attempted to directly regress ranking positions using a KL-divergence-based ranking loss. Given the predicted similarity scores and ground-truth rankings, we encouraged the probability distribution of predicted scores to match the true ranking distribution. 

The ranking loss is formally defined as:
\begin{equation}
\mathcal{L}_{\text{rank}} = \text{KL} \left( \text{softmax}(\mathbf{s}) \parallel \text{softmax}(\mathbf{r}) \right),
\end{equation}
where $\mathbf{s}$ represents the predicted similarity scores, and $\mathbf{r}$ represents the ground-truth ranking positions.

\noindent \textbf{Adaptive Weights Based on Query Similarity.}  
We explored an approach where the contribution of each neighboring feature was weighted by its cosine similarity to the query feature. Formally, the similarity between a neighboring feature $f_{\text{n}_j}$ and the query feature $f_q$ is computed as:
\begin{equation}
\text{sim}(f_{\text{n}_j}, f_q) = \frac{f_{\text{n}_j} \cdot f_q}{\|f_{\text{n}_j}\| \|f_q\| + \epsilon},
\end{equation}
where $\epsilon > 0$ is a small constant to avoid division by zero.

The final weight $w_j$ assigned to each neighboring feature is computed as:
\begin{equation}
w_j = \frac{\text{sim}(f_{\text{n}_j}, f_q)}{\sum_{k=1}^L \text{sim}(f_{\text{n}_k}, f_q) + \epsilon}.
\end{equation}

\begin{table}
\caption{\textbf{Performance comparison of Adaptive-MoF on different datasets.}}
\centering
\small
\begin{adjustbox}{width=\linewidth}
\begin{tabular}{clllll}
\toprule
Method & Pitts-30k-test & MSLS-Val & Pitts-250k-test & Nordland \\
\midrule
\multicolumn{5}{c}{\textbf{Fixed Parameters: L = 8}} \\
Baseline & 91.8 & 91.8 & 94.7 & 82.7 \\
5 & 92.59 & 85.41 & 95.63 & 82.9 \\
10 & 92.41 & 83.24 & 95.64 & 82.36 \\
25 & 92.34 & 82.03 & 95.69 & 81.98 \\
50 & 92.34 & 81.49 & 95.65 & 87.17 \\
\midrule
\multicolumn{5}{c}{\textbf{Fixed Parameters: L = 5}} \\
Baseline & 91.8 & 91.8 & 94.7 & 82.7 \\
5 & 92.55 & 89.05 & 95.46 & 84.09 \\
10 & 92.63 & 88.11 & 95.59 & 83.99 \\
25 & 92.66 & 87.43 & 95.68 & 83.82 \\
50 & 92.71 & 87.30 & 95.81 & 83.77 \\
100 & 92.72 & 86.62 & 95.79 & 83.73 \\
\bottomrule
\end{tabular}
\end{adjustbox}
\label{tab:adaptive_embodiedplace}
\end{table}

\subsection{Differences from SuperGlobal}

\begin{figure}
\centering
\includegraphics[width=\linewidth]{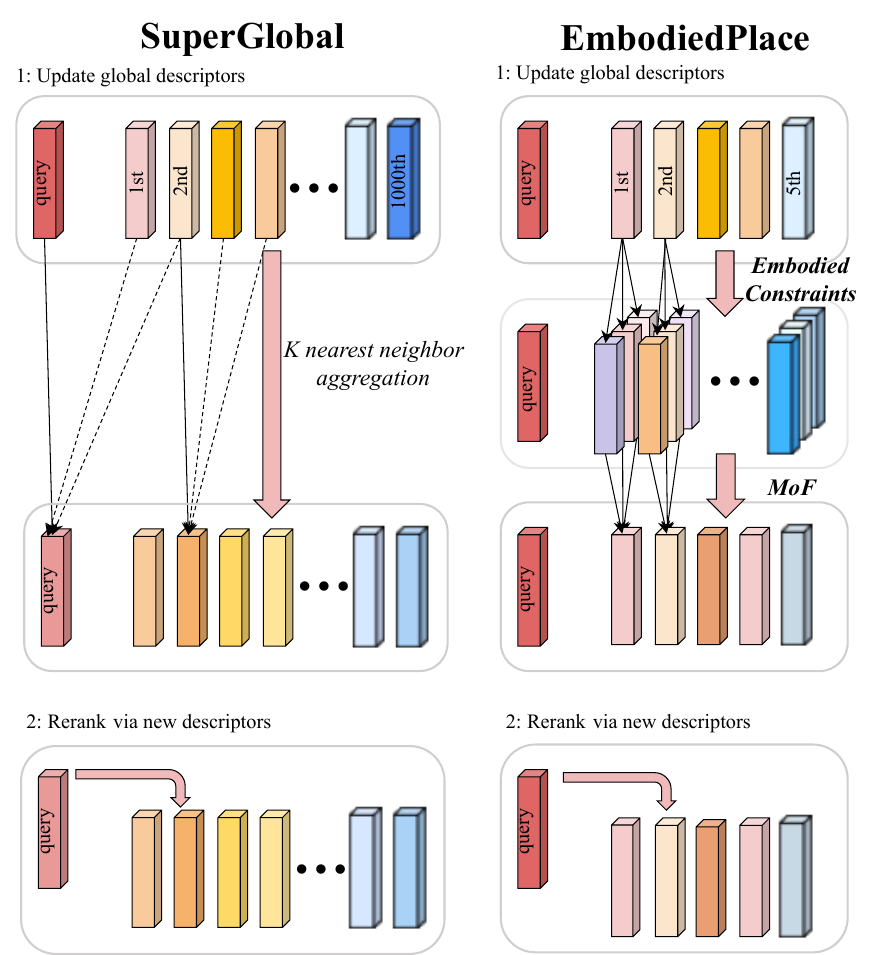}
\caption{\textbf{Differences between SuperGlobal and EmbodiedPlace.}}
\label{fig:superglobal}
\end{figure}

As illustrated in \cref{fig:superglobal}, EmbodiedPlace is inspired by SuperGlobal but differs in three key aspects: (1) instead of using KNN-based approximate search, we leverage embodied constraints in VPR to select candidate frames; (2) the weights are learned; and (3) we use a smaller re-ranking scope.

As shown in \cref{tab:superglobal_performance}, we conducted extensive parameter tuning experiments to compare and determine \textit{the optimal SuperGlobal configuration}.

\subsection{Which Embodied Constraint is the Best?}

\begin{figure}
\centering
\includegraphics[width=\linewidth]{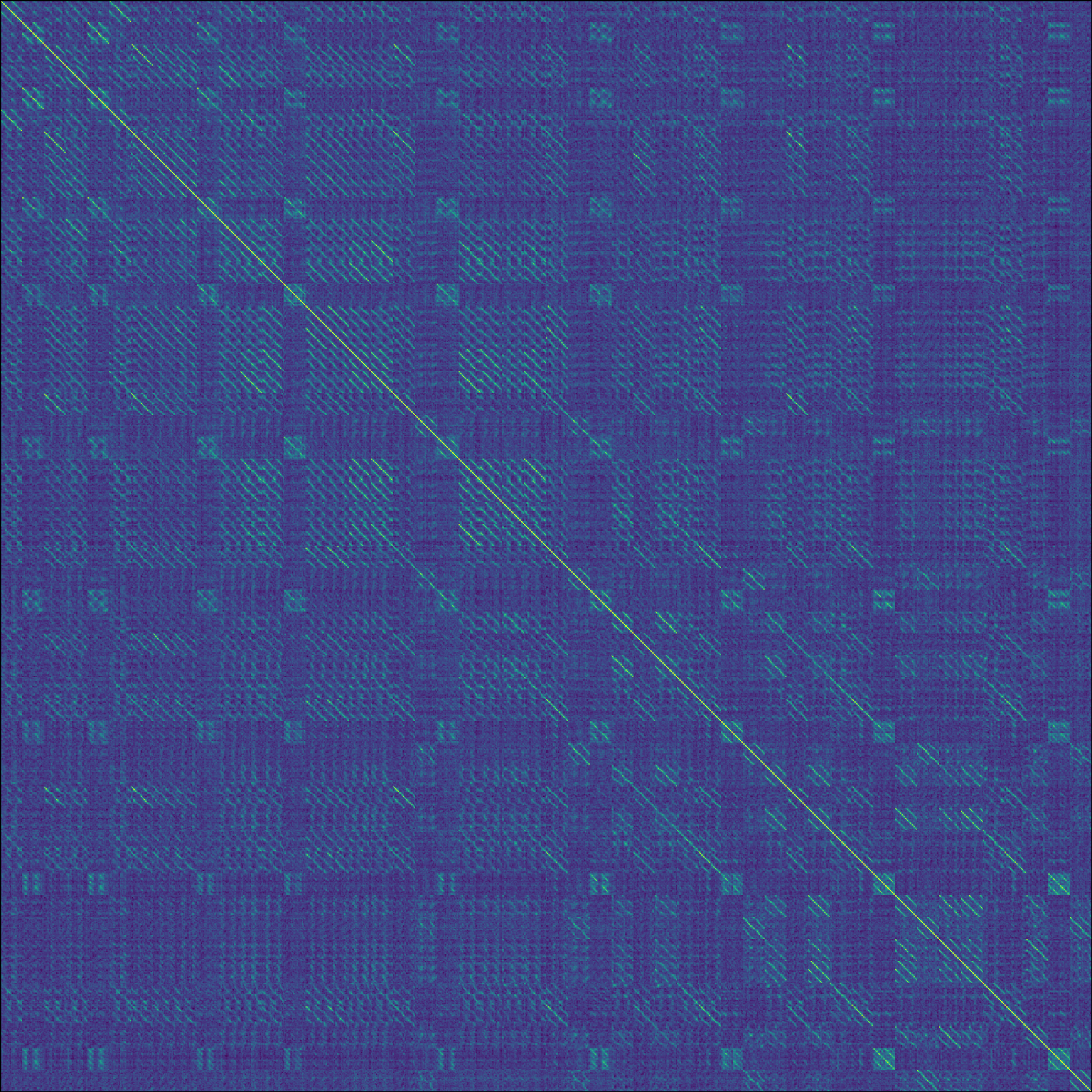}
\includegraphics[width=\linewidth]{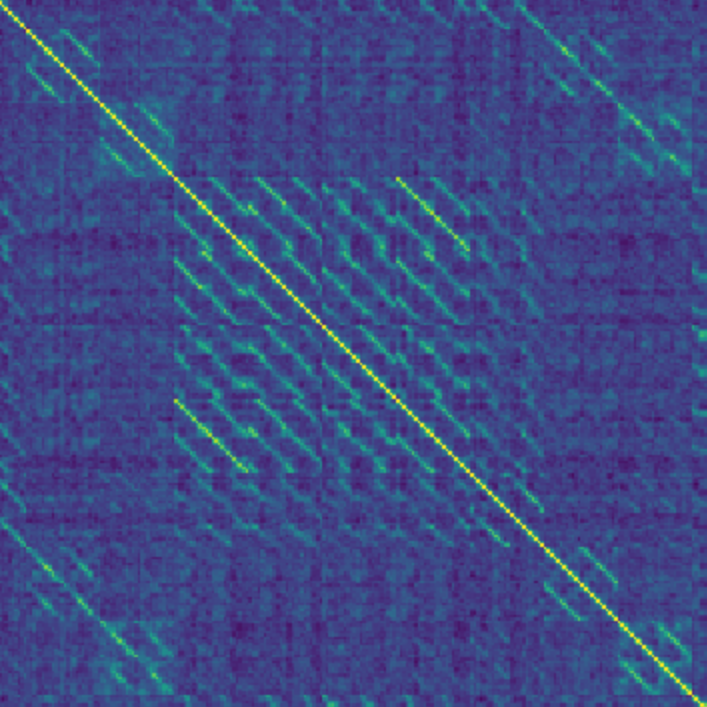}
\caption{\textbf{The global feature similarity matrix of the Pittsburgh-30k database.} It presents periodic similar blocks, with small blocks corresponding to the segmentation mode of panoramic camera images and large blocks corresponding to spatial correlations.}
\label{fig:sim_matrix}
\end{figure}

We consider four types of embodied constraints to adapt to different datasets, making direct comparison difficult. However, some intuitive observations can be made:

\begin{itemize}
    \item GPS tags and sequential information serve as strong embodied constraints, whereas self-similarity matrices of global features and local feature matching act as weaker (pseudo-embodied) constraints.
    \item The local feature matching is commonly used in visual localization systems, but it is challenging to construct for large-scale VPR datasets.
    \item The four embodied constraints may be used in combination, though redundancy may exist. As shown in \cref{fig:sim_matrix}, the self-similarity matrix of the Pitts dataset exhibits a clear periodic pattern, which corresponds to the GPS information and the splitting pattern of panoramic images.
\end{itemize}

\begin{table}
\caption{\textbf{Performance comparison of SuperGlobal on different datasets.}}
\centering
\small
\begin{adjustbox}{width=\linewidth}
\begin{tabular}{clllll}
\toprule
Method & Pitts-30k-test & MSLS-val & Pitts-250k-test & Nordland \\
\midrule
\multicolumn{5}{c}{\textbf{Fixed Parameters: $M = 100, \beta = 0.15$}} \\
Baseline & 91.8 & 91.8 & 94.7 & 82.7 \\
4 & 91.2 & 91.2 & 93.8 & 80.1 \\
6 & 90.6 & 90.5 & 93.5 & 78.7 \\
8 & 90.5 & 90.8 & 93.2 & 77.5 \\
10 & 90.2 & 90.4 & 93.1 & 76.5 \\
\midrule
\multicolumn{5}{c}{\textbf{Fixed Parameters: $M = 100, \beta = 0.25$}} \\
Baseline & 91.8 & 91.8 & 94.7 & 82.7 \\
4 & 90.9 & 91.6 & 93.6 & 80.2 \\
6 & 90.6 & 90.5 & 93.5 & 78.7 \\
8 & 90.4 & 90.9 & 93.2 & 77.4 \\
10 & 90.0 & 90.8 & 92.8 & 76.2 \\
\midrule
\multicolumn{5}{c}{\textbf{Fixed Parameters: $M = 400, \beta = 0.15$}} \\
Baseline & 91.8 & 91.8 & 94.7 & 82.7 \\
4 & 91.1 & 91.2 & 93.8 & 80.2 \\
6 & 90.6 & 90.3 & 93.4 & 78.6 \\
8 & 90.4 & 90.7 & 93.2 & 77.2 \\
10 & 90.0 & 90.7 & 92.8 & 76.1 \\
\bottomrule
\end{tabular}
\end{adjustbox}
\label{tab:superglobal_performance}
\end{table}

\subsection{Visualization of Results}
\begin{figure*}
\centering
\includegraphics[width=\textwidth]{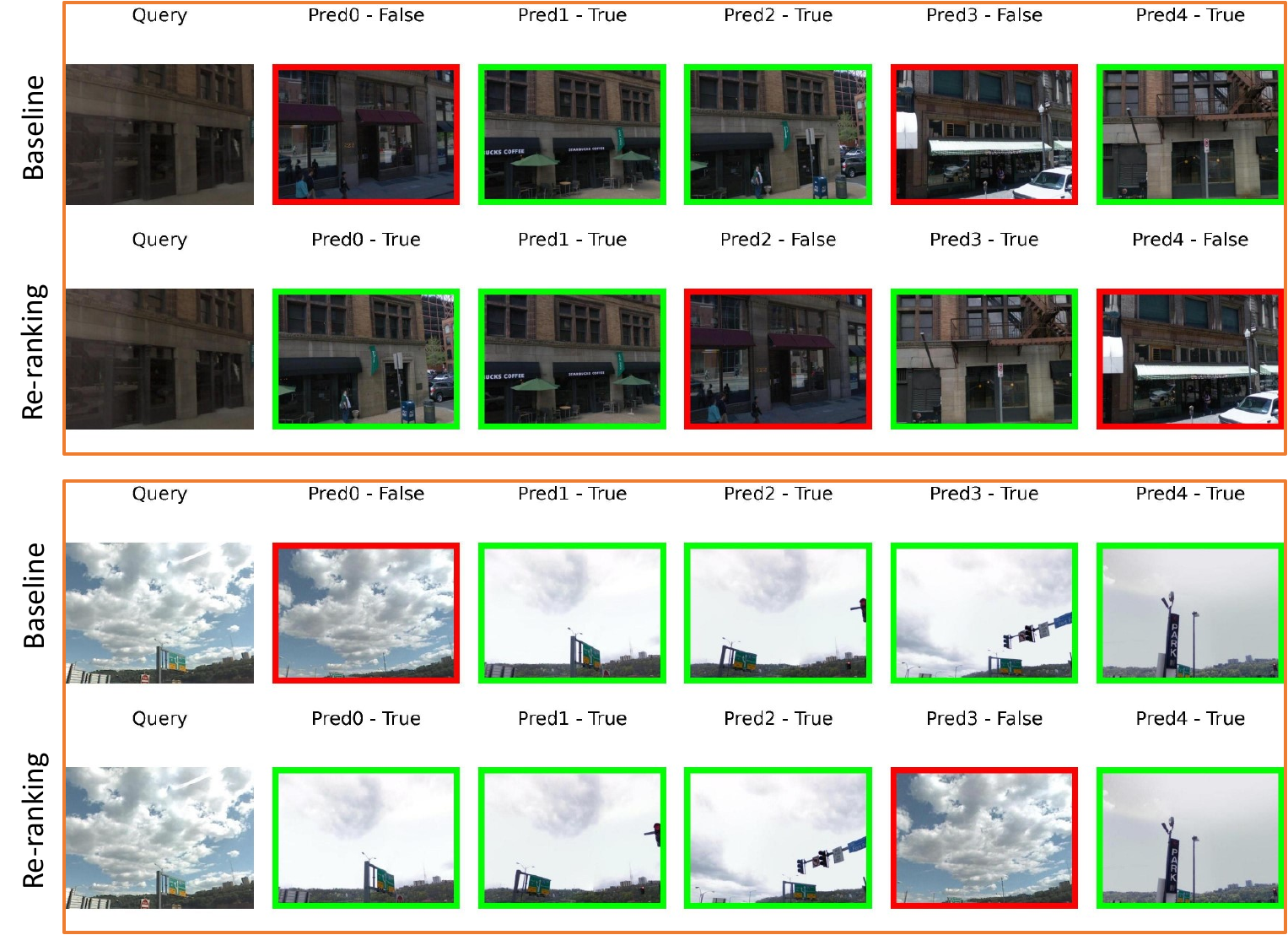}
\caption{\textbf{Visualization results of EmbodiedPlace on the Pittsburgh dataset.}}
\label{fig:visualization1}
\end{figure*}

\begin{figure*}
\centering
\includegraphics[width=\textwidth]{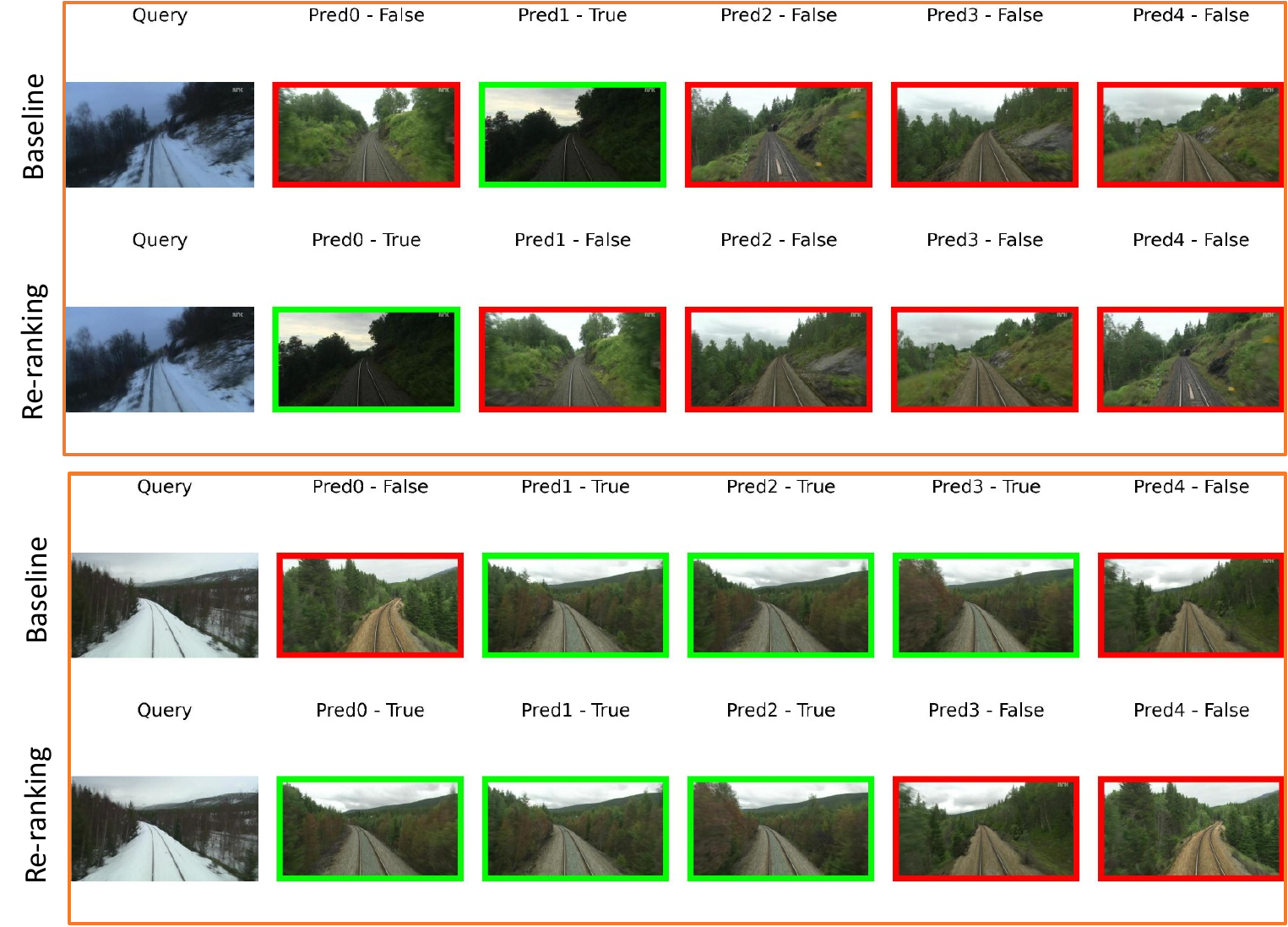}
\caption{\textbf{Visualization results of EmbodiedPlace on the Nordland dataset.}}
\label{fig:visualization2}
\end{figure*}

\begin{figure*}
\centering
\includegraphics[width=\textwidth]{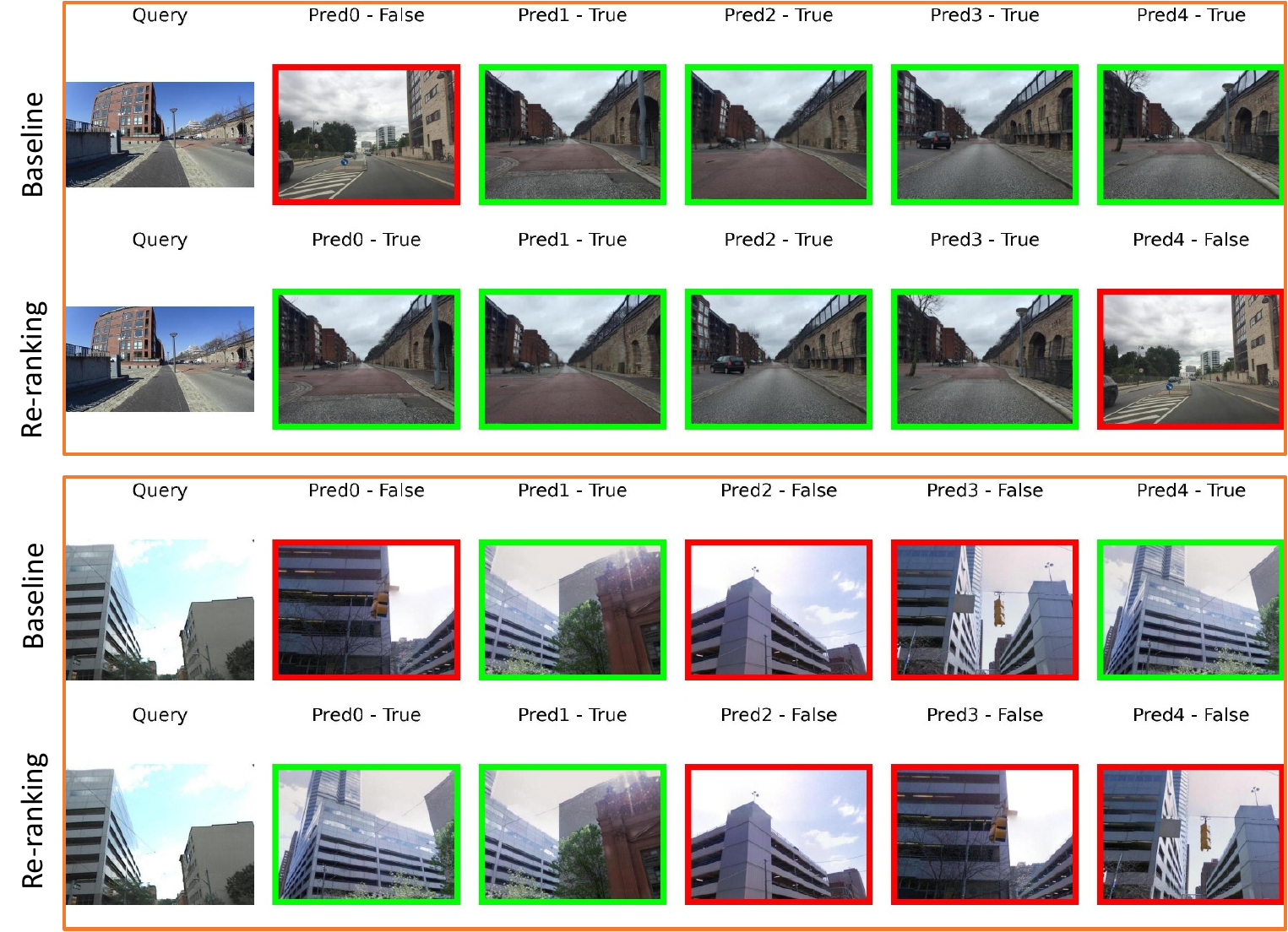}
\caption{\textbf{Visualization results of EmbodiedPlace on the MSLS dataset.}}
\label{fig:visualization3}
\end{figure*}

As shown in \cref{fig:visualization1}, \cref{fig:visualization2}, and \cref{fig:visualization3}, we visualize the query images along with their corresponding Top-5 retrieved images, demonstrating the performance improvement across different test sets.

\end{document}